\documentclass[letterpaper, 10 pt, conference]{ieeeconf}  
\usepackage[compress]{cite}
\usepackage[table]{xcolor}
\usepackage{booktabs,balance,subfigure}
\usepackage{graphicx}
\usepackage{amsmath}
\usepackage{epstopdf}
\usepackage{placeins}

\usepackage{multirow}

\IEEEoverridecommandlockouts                              

\title{\LARGE \bf
Lunar Terrain Relative Navigation Using a Convolutional Neural Network for Visual Crater Detection
}

\author{Lena M. Downes$^{1}$, Ted J. Steiner$^{2}$ and Jonathan P. How$^{3}$
\thanks{$^{1}$Lena M. Downes is with the Department of Aeronautics and Astronautics,
        Massachusetts Institute of Technology, Cambridge, MA 02139, USA, and 
        is a Draper Fellow with the Perception and Autonomy Group,
        Draper, Cambridge, MA 02139, USA
        {\tt\small lmdownes@mit.edu}}%
\thanks{$^{2}$Ted J. Steiner is with the Perception and Autonomy Group,
        Draper, Cambridge, MA 02139, USA
        {\tt\small tsteiner@draper.com}}%
\thanks{$^{3}$Jonathan P. How is with the Faculty of Aeronautics and Astronautics,
        Massachusetts Institute of Technology, Cambridge, MA 02139, USA
        {\tt\small jhow@mit.edu}}%
        \thanks{Research funded by Draper and computation support through Amazon Web Services}%
}

\begin{document}

\maketitle
\thispagestyle{empty}
\pagestyle{empty}

\begin{abstract}

Terrain relative navigation can improve the precision of a spacecraft's position estimate by detecting global features that act as supplementary measurements to correct for drift in the inertial navigation system. This paper presents a system that uses a convolutional neural network (CNN) and image processing methods to track the location of a simulated spacecraft with an extended Kalman filter (EKF). The CNN, called LunaNet, visually detects craters in the simulated camera frame and those detections are matched to known lunar craters in the region of the current estimated spacecraft position. These matched craters are treated as features that are tracked using the EKF. LunaNet enables more reliable position tracking over a simulated trajectory due to its greater robustness to changes in image brightness and more repeatable crater detections from frame to frame throughout a trajectory. LunaNet combined with an EKF produces a decrease of 60\% in the average final position estimation error and a decrease of 25\% in average final velocity estimation error compared to an EKF using an image processing-based crater detection method when tested on trajectories using images of standard brightness.
\end{abstract}

\section{INTRODUCTION}

Previous lunar missions have relied mostly on inertial navigation methods that time-integrate measurements from an inertial measurement unit (IMU) to estimate spacecraft position and velocity. As such, past lunar landing errors have been as high as several kilometers \cite{change} because inertial navigation systems drift and accumulate error over time. However, many of the locations of interest for future lunar landing missions are surrounded by or are close to hazardous terrain, motivating the requirement for increased estimation precision throughout the mission.  One method to reduce estimation error is terrain relative navigation (TRN) \cite{L07}. TRN obtains measurements of the terrain around a vehicle and uses those measurements to estimate the vehicle's position. A camera-based TRN system is an especially appealing option for space applications due to the availability and low cost of space-rated cameras, as well as the rich data they provide. 

Previous works on lunar TRN systems have developed crater detectors that are based upon traditional image processing methodologies like thresholding, edge detection, and filtering \cite{L12}. These crater detectors are highly sensitive to lighting conditions, noise, camera parameters, and viewing angles, and require precise modeling and specific tuning for different sets of images. Neural networks have been successfully applied as a robust solution to many computer vision problems due to their strength in generalizing from one dataset to another, but have rarely been used in space applications due to a perceived lack of confidence in their measurements \cite{Izzo2019}. We have previously developed a system that visually detects craters in a camera image using a neural network and matches those detected craters to a database of known lunar craters with absolute latitudes and longitudes \cite{scitech}. This crater detector is called LunaNet. This work explores the inclusion of LunaNet's measurements in a navigation system. LunaNet's repeatable detections from frame to frame across a trajectory and increased robustness to changes in camera lighting conditions enable an EKF that has increased precision position and velocity estimation with more reliable performance in variable lighting conditions.

\section{RELATED WORK}

TRN has been studied for years as a method to improve the landing accuracy of lunar landers \cite{L07,L06}. A common focus in TRN literature is to detect craters and use them as landmarks for localization. Many crater locations have already been catalogued since craters are present on nearly the entire lunar surface \cite{crater_coverage}. In addition, a detected crater's location can be uniquely identified based on the local constellation of nearby detected craters. 

Previous crater-based TRN systems utilized traditional image processing techniques to detect craters from visual imagery. However, these approaches are typically not robust to changes in lighting conditions, viewing angles, and camera parameters. Recent work has demonstrated that automated crater detection can achieve significant improvements in robustness by applying advances from the fields of computer vision and deep learning. CraterIDNet \cite{crateridnet} applied a fully convolutional network (FCN) \cite{fullyconv} to both visually detect craters in images and match those detected craters to known craters. An FCN is a type of convolutional neural network (CNN) that filters an image input using convolutional layers. Unlike our work, CraterIDNet focused on crater detection for the purpose of cataloguing craters, therefore it had low requirements for the precision of its crater detections. Our research focuses on repeatable detections over a trajectory for navigation, and for that reason we maintain strict requirements for precision of crater detection. Another recent system, DeepMoon \cite{deepmoon}, applied a CNN to detect craters from a digital elevation map (DEM) represented as overhead imagery. DeepMoon used a U-net \cite{unet} style architecture to perform pixel-wise classification of craters. DEM imagery has significantly different micro-scale variation than camera imagery because it is not affected by lighting effects such as glare and shadowing. In contrast, LunaNet uses camera images and thus must accommodate for shadows and other forms of visual noise, which introduces additional detection complexities. Moreover, DeepMoon did not focus on identifying or matching the craters it detected to known craters.

Crater identification, or matching of detected craters to known craters, has been explored in terms of two main types of problems. One problem is crater identification in a ``lost-in-space'' scenario, where no knowledge of spacecraft position is used to match detected craters to known craters. The other problem is crater identification with some information available about the spacecraft position, but without full confidence in that position information. This paper addresses on the latter problem, which \cite{L12} attempted to solve with a simple method: determining which craters would be in the camera field of view if the position information were correct, projecting these craters into image coordinates to obtain expected craters, and matching detected craters to the closest expected crater using least mean squares. This approach was successful if the position information provided was very close to the true position, but resulted in a large number of mismatches otherwise. An expansion upon the crater identification method of \cite{L12} was explored by \cite{Clerc}. This approach incorporated random sample consensus (RANSAC) \cite{ransac} to check whether the translation vectors between detected craters and known craters were consistent in images where multiple craters were detected. If a crater match produced a translation vector that was an outlier from the other crater matches, then that match was rejected. Both \cite{L12} and \cite{Clerc} followed a similar approach towards using the translation vector between the detected crater and the matched known crater in a navigation filter.

The system of \cite{L12} is used for a baseline comparison throughout this paper. This lunar terrain relative navigation system used a crater detector based on traditional image processing methods, which we refer to as the trinary edge detector due to its method of finding craters through trinary thresholding and edge detection. This work demonstrated the potential for craters to improve the state estimation of a spacecraft in lunar orbit. However, it struggled to consistently detect the same craters from frame to frame in a trajectory and experienced significant sensitivity to image quality, brightness, and shadowing. This paper explores how LunaNet, an improved, CNN-based crater detector developed in \cite{scitech}, can result in increased robustness of position and velocity estimation when used to generate features for an EKF.

 \section{Crater Detection}
This section serves to provide background on LunaNet, which was developed and analyzed in \cite{scitech}. LunaNet uses a CNN with U-net style architecture and traditional image processing methods to analyze camera imagery of the lunar surface simulated over orbital trajectories around the Moon. This architecture was based on the architecture of \cite{deepmoon}, called DeepMoon, which detected lunar craters from elevation imagery. A U-net \cite{unet} architecture was appealing because it is used for semantic segmentation, or detection and localization of distinct objects in an image, with pixel-level resolution. The crater detection problem is a semantic segmentation problem, where the objects that need to be detected and localized are craters. Low-level predictions of where the crater rims lie are needed in order to obtain precise crater detections. Semantic segmentation enables the labeling of each pixel in an image as crater, or not crater, and therefore provides high-precision labels of crater locations. 

LunaNet's CNN was initialized on the weights from DeepMoon and was additionally trained with 800 Lunar Reconnaissance Orbiter Camera (LROC) \cite{lro1,lro2,lro3,lro4} intensity images for ten epochs. Warm starting on \cite{deepmoon}'s neural network weights was appealing because their network was trained on many lunar elevation images. Although intensity imagery has a slightly different appearance than elevation imagery, additional training on intensity imagery could account for these differences. The training images were obtained from the LROC Wide Angle Camera (WAC) Global Morphology Mosaic with resolution of 100 meters per pixel. This mosaic is a simple cylindrical map projection comprising over 15,000 images from -90$^\circ$ to 90$^\circ$ latitude and all longitudes. The input training images are cropped from the mosaic using similar methods to those of DeepMoon: randomly cropping a square area of the mosaic, downsampling the image to 256$\times$256 pixels, and transforming the image to orthographic projection. Each cropped mosaic training image is paired with a binary image that has a black background and white rings corresponding to the pixel locations of known crater rims in the cropped mosaic image. 

LunaNet's CNN outputs a grayscale image with bright values indicating pixels that are predicted to be part of a crater rim.  This image is processed to identify discrete crater detections in the image. The CNN output is first thresholded by intensity to obtain predictions with the highest level of detection certainty. Predictions with greater than 90\% certainty are set to intensity of 255, and predictions with less than that are set to 0. The predictions are then eroded to find single-pixel lines of crater rim detections. Following this, the contours of the eroded predictions are found and these are used to eliminate individual detections which are smaller than three pixels. These contours are then fit with ellipses, with 15\% being the maximum ratio of ellipse minor axis to major axis. The ellipses are not allowed to be more elliptical because true craters are generally circular, and detections that are more elliptical tend to be false detections. This process produces the final crater detections outputs of LunaNet for the image: discrete, closed, circular ellipses that represent the rims of the detected craters. Analysis in \cite{scitech} demonstrates the high level of robustness of LunaNet to noise and variations in image brightness. 

\section{Crater Identification}
The spacecraft position estimate from the EKF enables the prediction of which known craters from the two crater databases (the 5-20 km database from \cite{crater_small} and the $>$20 km database from \cite{crater_big}) are expected to be visible in the camera frame. The estimated latitude and longitude limits of the image are used to discard all database craters which do not fit into the expected image field of view. The remaining database craters are then transformed into image coordinates. These expected craters are the craters that are expected to be in the camera frame based off of the position estimate and the a priori knowledge of craters in that location. The detected craters are then matched to the expected craters through a process that combines the methods of \cite{L12} and \cite{Clerc}. Each detected crater is paired with an expected crater according to a least mean squares fit, which takes into account the crater center coordinates and the approximate diameters of the detected and expected craters. The vectors between pairs of expected and detected craters are processed with affine transform RANSAC to guard against false matches \cite{ransac}. Vectors which are classified as outliers by RANSAC are discarded, and only the crater pairs with inlier vectors are accepted as crater matches. 

\section{Extended Kalman Filter}
A typical feature-based extended Kalman filter (EKF) like the one used in \cite{L12} is utilized to localize the simulated spacecraft. The state vector of the EKF is written as
\begin{equation}\label{eq:state}
X(k) = [V_{c} \: X_{c} \: X_{F1} \: X_{F2}\: \ldots \: X_{FN}]^{T} 
\end{equation}
where $V_{c}$ is the velocity of the camera in lunar-centered lunar-fixed coordinates (LCLF), $X_{c}$ is the position of the camera in LCLF, and $X_{Fi}$ is the position of the $i^{th}$ feature in LCLF. The state propagation equations are linear approximations in the following equations since the motion can be considered planar for short segments of orbits,
\begin{align}\label{eq:vel}
 V_{c}(k) &= V_{c}(k-1)+U_{t}(k) dT , \\
\label{eq:pos}
 X_{c}(k) &= X_{c}(k-1)+ V_{c}(k-1) dT +U_{t}(k) \frac{dT^{2}}{2} ,
\end{align}
where $U_{t}(k)$ is the acceleration that the spacecraft experiences at time $k$ in LCLF, $dT$ is the time step, and $X_{c}(k)$ is the position of the camera at time $k$ in LCLF. The detected craters which have been successfully matched to database craters are used to initialize features. The longitude and latitude from the database crater is transformed to LCLF and entered as a feature in the state vector according to \eqref{eq:state}. As these same craters are detected again in subsequent frames, $z_{Fi}$, the unit vector from the $i^{th}$ feature to the current camera position, is calculated as 
\begin{equation}\label{eq:vec}
 z_{Fi}(k) = \frac{X_{Fi}(k)-X_{c}(k)}{\| X_{Fi}(k)-X_{c}(k) \|}.
\end{equation}
The residual for feature $i$, $\tilde{y}_{Fi}(k)$, is calculated as 
\begin{equation}\label{eq:res}
 \tilde{y}_{Fi}(k) = z_{Fi}(k)-\hat{z}_{Fi}(k|k-1),
\end{equation}
and produces the vector between the detected crater location and the expected crater location, based on propagation. The complete residual $\tilde{y}(k)$ consists of the feature residuals and the camera residuals. The complete residual is used in the state update using the typical Kalman gain calculation
\begin{align}\label{eq:k}
 K(k) &= P(k|k-1)H(k)^{T}S(k)^{-1} \\
\label{eq:kal}
 X(k|k) &= X(k|k-1)+K(k) \tilde{y}(k)
\end{align}
where $K$ is the Kalman gain, $P$ is the covariance estimate, $H$ is the observation matrix, and $S$ is the residual covariance. Every time a previously initialized feature is seen again, its feature residual, \eqref{eq:res}, is used in the state update. If an initialized feature is not seen again then its feature residual is not calculated or used to produce the state update, but the state estimates $X_{Fi}$ of all initialized feature vectors are maintained in the state vector and updated in case any of those features are seen again.

 \section{Simulation Results}
 More persistent feature tracking, in which a feature is initially detected and is then repeatedly detected in future frames, improves accuracy of the state estimate in TRN since longer tracks correspond to higher signal to noise ratios \cite{JPLTRN}. The trinary edge detector struggles to persistently detect the same craters in subsequent frames. This contributes to the shorter length of crater detection tracks in each trajectory for the trinary edge detector, as characterized in Table~\ref{table_lendet}. LunaNet tracks craters on average for twice as many frames as the trinary edge detector does. Each subsequent detection of a crater after it is initialized increases the confidence of the feature state estimate, contributing to the accuracy of the vehicle state estimate. The inconsistency in feature detection of the trinary edge detector also results in more false matches, where detected craters are mistakenly matched to the wrong database craters. These false matches are usually thrown out by RANSAC in the crater matching system, but occasionally they are accepted as good matches. These occurrences contribute to the shorter average tracking length, since false matches could be matched to different database craters in consecutive frames, thereby causing the EKF to diverge.

\begin{figure*}[t!]
  \centering
  \subfigure[Using trinary edge crater detector from \cite{L12}. Standard imagery. \label{fig:det_singh}]{\includegraphics[trim=10 5 5 5,clip,width=.48\columnwidth]{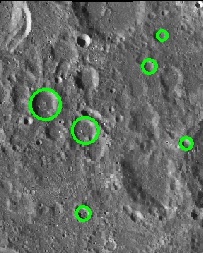}}
  \hfill
  \subfigure[Using LunaNet. Standard imagery. \label{fig:det_luna}]{\includegraphics[trim=10 5 5 5,clip, width=.48\columnwidth]{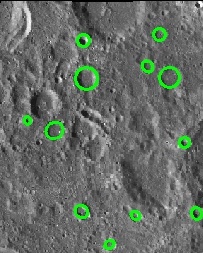}}
  \hfill
  \subfigure[Using trinary edge crater detector from \cite{L12}. Image brightness increased by 30\%. \label{fig:det_singh_b}]{\includegraphics[trim=10 5 5 5,clip,width=.48\columnwidth]{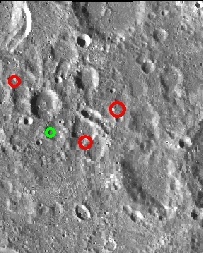}}
  \hfill
  \subfigure[Using LunaNet. Image brightness increased by 30\%. \label{fig:det_luna_b}]{\includegraphics[trim=10 5 5 5,clip, width=.48\columnwidth]{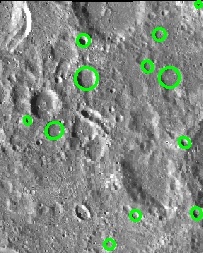}}
  \caption{Comparison of the trinary edge detector vs. LunaNet on a standard image and one with 30\% increased brightness. On the standard image, LunaNet detects more craters than the trinary edge detector. On the brighter image, LunaNet detects all but one of the same craters as in the standard image. In contrast, the trinary edge detector detects only one of the same craters as in the standard image. It also produces three false matches, marked in red. These false matches were found by human inspection, and would be accepted as true detections by the EKF.}
  \label{fig:det}
  \end{figure*}

Another key improvement of LunaNet is that it is less sensitive to changes in brightness levels in the image. We have demonstrated LunaNet's robustness by training it on the same set of images that the trinary edge detector was tuned for, and then testing it and the trinary edge detector over trajectories of images 30\% brighter and 30\% darker than the training and tuning set. A 30\% difference in brightness level could unknowingly occur due to differences in time of day, exposure of the camera, or albedo properties of the local lunar surface. Fig. \ref{fig:det} shows the performance of LunaNet and the trinary edge detector on a representative image of standard brightness and higher brightness. LunaNet detects almost all of the same craters in both images, and human inspection revealed all of them to be true matches. 

The trinary edge detector experiences a sharp degradation in performance on the brighter imagery, as seen in Fig. \ref{fig:det_singh_b}. The trinary edge detector detects completely different craters than it did in the standard image, and human inspection revealed that only one of the accepted and matched detections was properly matched to a known crater. This demonstrates the trinary edge detector's lower level of robustness to image variation,
and reveals a concerning propensity towards accepting incorrectly matched craters as true detections. These false matches are fed into the EKF and result in error spikes, drifting, and general degradation of the EKF estimate. LunaNet has more consistent performance than the trinary edge detector regardless of the image brightness. This quality is extremely important in the context of a mission because it is difficult to determine a priori exactly how the lunar surface will appear in a camera image. With the trinary edge detector, the qualities of the image would need to be known with a high level of confidence before the mission in order to ensure that the crater detector could be properly tuned. LunaNet does not require tuning for image brightness.

A higher number of detected features corresponds to lower position estimate uncertainty in terrain relative navigation, as evinced by \cite{tedlandmark}. Table~\ref{table_numdet} shows numerical measures of how LunaNet outperforms the trinary edge detector in average number of craters detected, with LunaNet detecting 18 craters on average over a trajectory, where the trinary edge detector detects 16 craters on average. Although the trinary edge detector detects a higher number of craters with changes in image brightness, visual inspection revealed that a greater percentage of these detections were incorrectly matched to the wrong known craters. These higher numbers of false matches cause the EKF to diverge and result in less reliable state estimation.

\begin{figure*}[t!]
  \centering
  \subfigure[Using the trinary edge detector, based on crater detector in \cite{L12}.\label{fig:pos_singh}]{\includegraphics[width=.94\columnwidth]{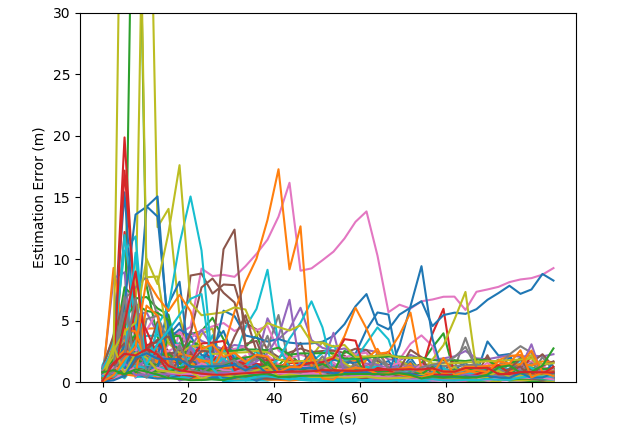}}
  \hfill
  \subfigure[Using LunaNet detections. \label{fig:pos_luna}]{\includegraphics[width=.94\columnwidth]{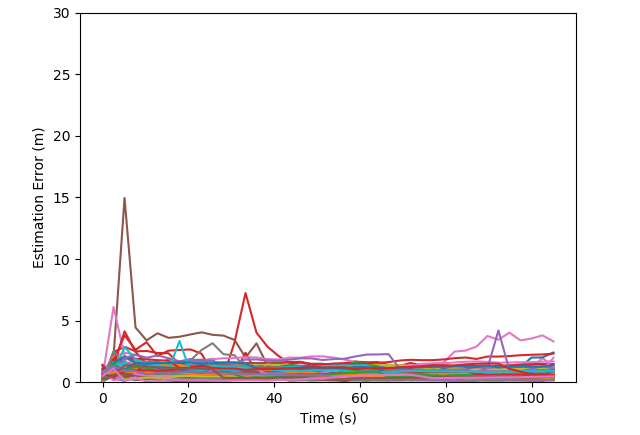}}
  \caption{The norm of EKF position error for a Monte-Carlo simulation of 100 different lunar orbit trajectories.}
  \label{fig:pos}
\end{figure*}

LunaNet's persistent feature tracking and large number of true crater detections, regardless of changes in lighting conditions, contribute to an EKF that converges more rapidly to accurate position and velocity estimates. In these simulations the EKF is initialized on the true position and velocity information, and then the position and velocity are propagated using the true acceleration with a simulated random noise on the order of 0.1 m/s\textsuperscript{2}, and a time step of 2.5 seconds between updates. Fig.~\ref{fig:pos_luna} shows the norm of the position estimation error over 100 different trajectories for a LunaNet-based EKF on images with standard brightness. The EKF using LunaNet experiences fewer spikes in estimation error and converges more quickly than the EKF using the trinary edge detector in Fig.~\ref{fig:pos_singh}, which shows the norm of the position estimation error over the same 100 trajectories.   The final average estimation errors using LunaNet are 0.53 m and 0.09 m/s. The final average estimation error of the trinary edge detector is 1.33 m and 0.12 m/s. LunaNet also results in more consistent performance, as demonstrated by its less frequent estimation error spikes.

\begin{table}[h]
\caption{Average Number of Times a Crater is Detected}
\vspace*{-.15in}
\label{table_lendet}
\begin{center}
\begin{tabular}{|cl||c|}
\hline
\multicolumn{2}{|c||}{\bf Image and Crater}  & \bf Average Length  \\
\multicolumn{2}{|c||}{\bf Detector Type} & \bf of Detection \\\hline
\multirow{2}{*}{Standard Imagery} & LunaNet & 13\\
& Trinary Edge Detector & 5 \\\hline
\multirow{2}{*}{Bright Imagery} & LunaNet & 13\\
& Trinary Edge Detector & 6 \\\hline
\multirow{2}{*}{Dark Imagery} & LunaNet & 13\\
& Trinary Edge Detector & 7 \\
\hline
\end{tabular}
\end{center}

\caption{Average Number of Craters Detected}
\vspace*{-.15in}
\label{table_numdet}
\begin{center}
\begin{tabular}{|cl||c|}
\hline
\multicolumn{2}{|c||}{\bf Image and Crater}  & \bf Average Number   \\
\multicolumn{2}{|c||}{\bf  Detector Type} & \bf  of Craters  \\\hline
\multirow{2}{*}{Standard Imagery} & LunaNet & 18\\
& Trinary Edge Detector & 16 \\\hline
\multirow{2}{*}{Bright Imagery} & LunaNet & 19 \\
& Trinary Edge Detector & 29 \\\hline
\multirow{2}{*}{Dark Imagery} & LunaNet & 16 \\
& Trinary Edge Detector & 23 \\\hline
\end{tabular}
\end{center}
\end{table}

\begin{figure*}[t!]
  \centering
  \subfigure[Using the trinary edge detector, based on crater detector in \cite{L12}.\label{fig:vel_singh}]{\includegraphics[width=.94\columnwidth]{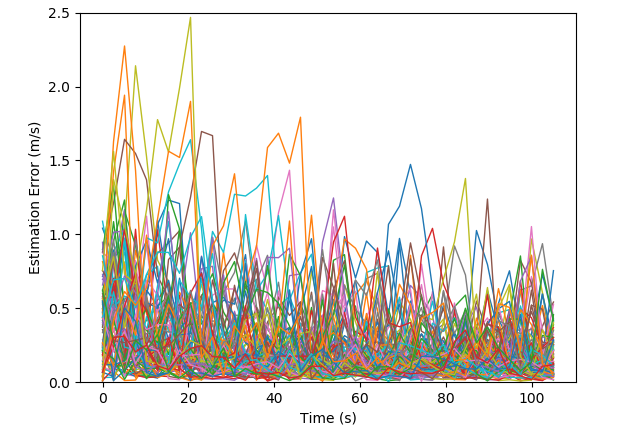}}
  \hfill
  \subfigure[Using LunaNet detections. \label{fig:vel_luna}]{\includegraphics[width=.94\columnwidth]{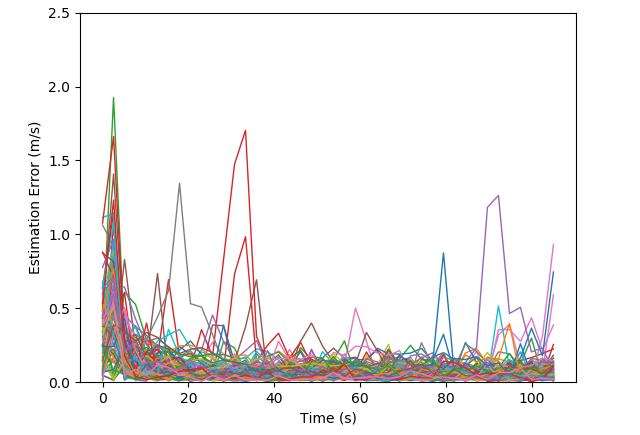}}
  \caption{The norm of EKF velocity error for a Monte-Carlo simulation of 100 different lunar orbit trajectories.}
  \label{fig:vel}
\end{figure*}
\FloatBarrier


In contrast, the trinary edge detector in Fig.~\ref{fig:pos_singh} has many trajectories with estimation errors that spike, diverge and drift for large portions of the simulation. The same comparison is evident between the velocity estimation errors in Fig.~\ref{fig:vel}. The trinary edge detector again has far more error spikes, likely due to its less persistent detection of craters. The trinary edge detector's estimation error indicates that this method is less reliable and more prone to false detections, false matches, and overall lower numbers of detections.

\begin{figure*}[t!]
  \centering
  \subfigure[Position errors\label{fig:pos_all}]{\includegraphics[width=.94\columnwidth]{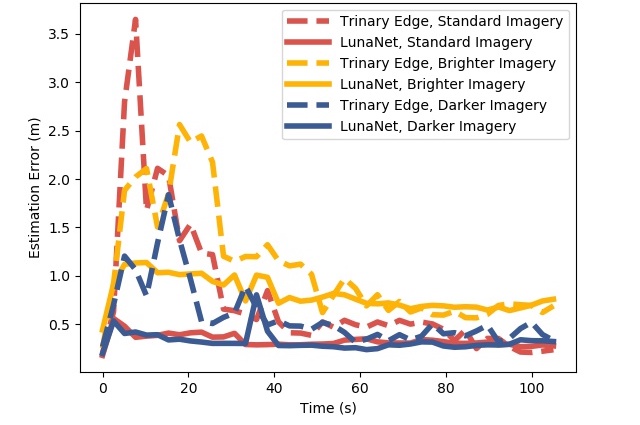}}
  \hfill
  \subfigure[Velocity errors\label{fig:vel_all}]{\includegraphics[width=.94\columnwidth]{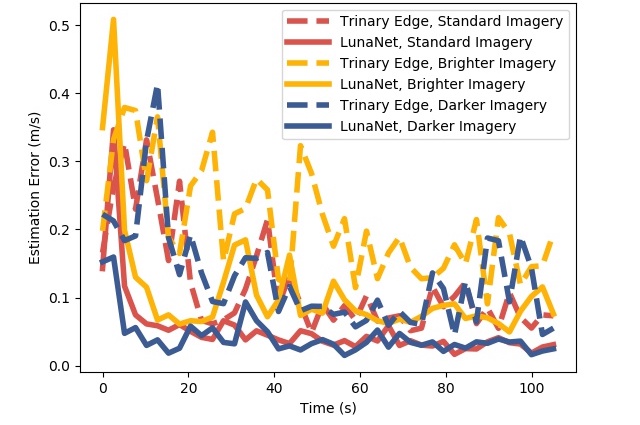}}
  \caption{One sigma errors of the norms of EKF errors for a Monte-Carlo simulation of 20 different lunar orbit trajectories. EKFs were run on a set of images similar to those that the detectors were trained or tuned on, on images 30\% brighter, and on images 30\% darker.}
\end{figure*}

A comparison of the EKF performance of LunaNet and the trinary edge detector with standard imagery, brighter imagery, and darker imagery is shown in Figures \ref{fig:pos_all} and \ref{fig:vel_all}. These figures show the average estimation error of 20 trajectories from Monte-Carlo simulations like those in Figs. \ref{fig:pos} and \ref{fig:vel}. LunaNet converges faster and experiences a smaller degradation in performance than the trinary edge detector, even on images that are brighter or darker than those that it was trained on. Overall, LunaNet results in fewer estimation error spikes and faster convergence on images of different brightness levels when compared to the performance of the trinary edge detector on the same images and trajectories.
\section{Conclusions}

In summary, this work improved an existing lunar terrain relative navigation system by replacing the crater detector with LunaNet, a neural network-based crater detector developed in \cite{scitech}. The trinary edge detector struggles to repeatedly detect craters in subsequent frames after initialization, while LunaNet persistently detects craters from frame to frame. LunaNet achieves more consistently long feature tracks, meaning that once a crater is initialized, it is repeatedly detected in subsequent frames more times than the trinary edge detector does. This quality causes estimation uncertainty to decrease more rapidly, resulting in a more reliable position estimate. The trinary edge detector is highly tuned to images with specific brightness levels, while LunaNet does not require additional training to perform comparably well on images with different levels of brightness. With a 30\% brightness increase or decrease compared to its training images, LunaNet detected the same average number of craters over twenty trajectories. In contrast, the trinary edge detector experienced a degradation in performance and detected a higher number of false crater matches. When used with an EKF, LunaNet enables improved localization performance compared to the trinary edge detector. A LunaNet-based EKF obtains higher accuracy and converges more quickly than an EKF based on a traditional crater detector. A LunaNet-based EKF obtains a decrease of 60\% in the average final position estimation error and a decrease of 25\% in the average final velocity estimation error for standard brightness images when compared to an EKF based on the trinary edge detector. 

\bibliographystyle{IEEEtran}
\bibliography{root}

\end{document}